\documentclass[10pt,twocolumn,letterpaper]{article}

\usepackage[pagenumbers]{utils/cvpr}      

\usepackage{graphicx}
\usepackage{amsmath}
\usepackage{amssymb}
\usepackage{booktabs}
\usepackage{comment}
\usepackage{multirow}
\usepackage{tabularx}
\usepackage{float}
\usepackage{xcolor}
\usepackage{bbm}
\usepackage{import}
\usepackage{tikz}
\usepackage{comment}

\usepackage{pifont} 
\usepackage{booktabs}
\usepackage{multirow}
\usepackage{adjustbox}
\usepackage{fontawesome} 
\usepackage{amsmath} 

\definecolor{opA}{rgb}{0.9,0.6,0.0}
\definecolor{opB}{rgb}{0.35,0.70,0.90}
\definecolor{opC}{rgb}{0.8,0.40,0.0}
\definecolor{opD}{rgb}{0.0,0.60,0.50} %
\definecolor{opE}{rgb}{0.8,0.6,0.7}
\definecolor{opF}{rgb}{0.,0.45,0.70} 

\definecolor{pltBlue}{rgb}{0.12156862745098039, 0.4666666666666667, 0.7058823529411765}
\definecolor{pltOrange}{rgb}{1.0, 0.4980392156862745, 0.054901960784313725}
\definecolor{pltGreen}{rgb}{0.17254901960784313, 0.6274509803921569, 0.17254901960784313}
\definecolor{pltRed}{rgb}{0.8392156862745098, 0.15294117647058825, 0.1568627450980392}
\definecolor{pltViolet}{rgb}{0.5803921568627451, 0.403921568627451, 0.7411764705882353}
\definecolor{pltBrown}{rgb}{0.5490196078431373, 0.33725490196078434, 0.29411764705882354}
\definecolor{pltMagenta}{rgb}{0.8901960784313725, 0.4666666666666667, 0.7607843137254902}
\definecolor{pltGray}{rgb}{0.4980392156862745, 0.4980392156862745, 0.4980392156862745}
\definecolor{pltLightGreen}{rgb}{0.7372549019607844, 0.7411764705882353, 0.13333333333333333}
\definecolor{pltCyan}{rgb}{0.09019607843137255, 0.7450980392156863, 0.8117647058823529}
\definecolor{pltPink}{rgb}{0.49803921568627, 0.49803921568627, 0.49803921568627}

\definecolor{cmarkcolor}{rgb}{0.49,0.74,0.49}
\definecolor{xmarkcolor}{rgb}{0.86,0.34,0.34}

\newcommand{\xmark}{\textcolor{xmarkcolor}{\ding{55}}}

\def\link#1{
    \ifx&#1&
        \xmark{}
    \else
        {\href{#1}{\faExternalLink}}
    \fi
}

\usepackage{siunitx}
\usepackage{mathrsfs}
\definecolor{cvprblue}{rgb}{0.21,0.49,0.74}
\usepackage[pagebackref,breaklinks,colorlinks,citecolor=cvprblue]{hyperref}

\usepackage[capitalize]{cleveref}
\usepackage{layouts}
\crefname{section}{Sec.}{Secs.}
\Crefname{section}{Section}{Sections}
\Crefname{table}{Table}{Tables}
\crefname{table}{Tab.}{Tabs.}

\begin{document}

\title{3DiFACE: Diffusion-based Speech-driven 3D Facial Animation and Editing}

\author{
Balamurugan Thambiraja\textsuperscript{1}
\quad 
Sadegh Aliakbarian\textsuperscript{3}
\quad 
Darren Cosker\textsuperscript{3}
\quad
Justus Thies\textsuperscript{1,2}
\\
\\
\textsuperscript{1} Max Planck Institute for Intelligent Systems, Tübingen, Germany \\
\textsuperscript{2} Technical University of Darmstadt  \quad  \textsuperscript{3} Microsoft Mixed Reality \& AI Lab \\
\url{https://balamuruganthambiraja.github.io/3DiFACE}
}

\twocolumn[{%
\renewcommand\twocolumn[1][]{#1}%
\maketitle
\begin{center}
    \centering
    \captionsetup{type=figure}
    \vspace{-0.5cm}
    \includegraphics[width=\textwidth]{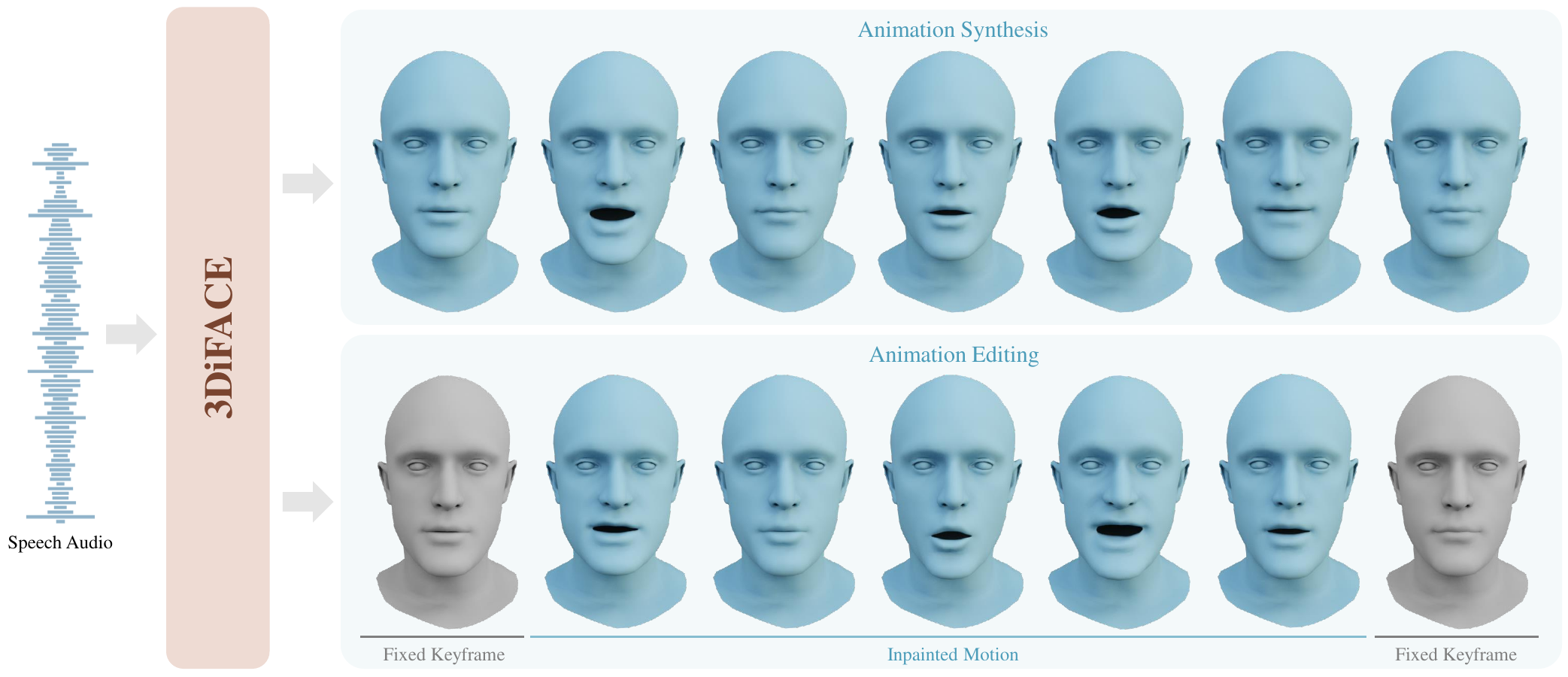}
      \caption{
      \textit{3DiFACE} is a novel diffusion-based method for speech-driven 3D facial animation.
      Given an audio sequence, our method can generate motion sequences with lip-sync and stochasticity. 
      Additionally, 3DiFACE can be used for audio-consistent motion editing.
      }
      \vspace{0.2cm}
      \label{fig:teaser}
\end{center}
}]

\begin{abstract}
We present 3DiFACE, a novel method for personalized speech-driven 3D facial animation and editing.
While existing methods deterministically predict facial animations from speech, they overlook the inherent one-to-many relationship between speech and facial expressions, i.e., there are multiple reasonable facial expression animations matching an audio input.
It is especially important in content creation to be able to modify generated motion or to specify keyframes.
To enable stochasticity as well as motion editing, we propose a lightweight audio-conditioned diffusion model for 3D facial motion.
This diffusion model can be trained on a small 3D motion dataset, maintaining expressive lip motion output.
In addition, it can be finetuned for specific subjects, requiring only a short video of the person.
Through quantitative and qualitative evaluations, we show that our method outperforms existing state-of-the-art techniques and yields speech-driven animations with greater fidelity and diversity.

\vspace{-0.5cm}
\end{abstract}

\section{Introduction}\label{sec:intro}
Generating faithful 3D animations is crucial for realistic and immersive digital experience in games, movies, and other human-centric entertainment applications.
Earlier works on 3D facial animation focused on animating faces based on procedural rules that map audio features to facial animation parameters \cite{cohen01_avsp, edwards2016jali}.
With advancements in machine learning, data-driven methods have been widely used to generate animations conditioned on audio input \cite{zhang20213d, peng2023emotalk, cudeiro2019capture, meshtalk, fan2022faceformer, xing2023codetalker}.
Despite the recent advances in 3D facial animation, most of the existing methods learn a deterministic mapping between audio and facial animation, overlooking the inherent one-to-many relationship—one audio signal can match many different facial expression animations. 
This limits the diversity of synthesized animations.
In addition, most of the existing works focus solely on motion synthesis, but do not address motion editing, e.g. inbetweening facial motion between two keyframes.
However, motion editing is an equally interesting and challenging problem with high relevance for practical content creation applications in the gaming and movie industry.
In this work, we address this gap and propose a diffusion-based architecture to perform speech-driven motion synthesis and editing.
In doing so, we face two challenges: 
(i) Diffusion models are known to require large training sets \cite{rombach2022high}, yet the size of existing high-quality speech-to-3D-animation datasets is limited.
(ii) Facial movements are highly person-specific. This requires the learning of person-specific speaking styles into the synthesis and editing pipeline.  
Especially for facial motion editing applications, without person-specific speaking-style, we observe an abrupt change in speaking style between the edited and unedited motion, resulting in unrealistic animations.

Recent works such as EMOTE~\cite{EMOTE} and DiffPoseTalk~\cite{sun2023diffposetalk} employ head trackers to annotate large-scale-video datasets with pseudo ground truth data and train their models on the resulting dataset.
While effectively solving data scarcity, the synthesis fidelity is limited by the quality of the trackers and it is inferior to models that were trained on smaller datasets with higher quality \cite{voca}. 
Imitator~\cite{imitator} is a method that addresses speaker style adaptation, by learning a person-specific motion decoder, however, the motion is deterministic and does not allow for motion editing.

In this work, we propose a diffusion-based pipeline for speech-driven facial animation which can be efficiently trained on small datasets and enabling us to optimize person-specific speaking styles. 
In contrast to previous works~\cite{faceformer,imitator,EMOTE,codetalker} that rely on transformer-based architectures~\cite{vaswani2017attention}, we propose an audio-conditioned diffusion model using a 1D convolutional backbone.
In addition, we employ a concatenation-based convolution block over the widely used attention mechanism to inject audio conditioning signals into the 1D convolutional model and we propose a window-based training scheme.
Based on our light-weight architecture, we take inspiration from Imitator~\cite{imitator} and propose a person-specific fine-tuning technique, that only requires a short video of the target actor. 
The diffusion architecture as well as the personalization allows us to generate and edit animations with person-specific speaking style.
Since we regress vertex displacements directly, our method can capture more subtle speaking style variations than approaches that regress coefficients of a parametric head model \cite{flame}.
Through qualitative and quantitative studies, we show that our proposed pipeline outperforms the state of the art in synthesizing and editing realistic 3D facial animations conditioned on input audio, while preserving the speaking style of a target subject. 
In summary, our contributions are:
\begin{itemize}
  \item We introduce a novel diffusion-based method for speech-driven 3D facial animation that can be efficiently trained on small-scale high-quality datasets to generate diverse animations from a single audio source.
  \item Our method captures person-specific speaking styles from short reference videos, enabling personalized facial animations.
  \item We demonstrate exciting applications in editing facial motions, such as seamless motion interpolation, keyframing, and unconditional facial motion synthesis.
\end{itemize}
\section{Related Work}
\label{sec:related}
Numerous prior investigations have delved into the realm of speech-driven generation, with a predominant emphasis on synthesizing 2D talking head videos.
However, especially for 3D content creation in games, movies, and immersive telepresence, speech-driven 3D facial animation is of high interest to the research community.
Our work generates 3D facial animations using a denoising diffusion probabilistic model, therefore we review motion diffusion work.
\vspace{-0.1cm}
\paragraph{Talking Head Videos}
Approaches for talking head video generation can be mainly categorized into two groups: 
directly generating RGB videos from speech on the one hand, and
utilizing a 3D Morphable Model (3DMM) for guided 2D or 3D rendering on the other.
Suwajanakorn et al.~\cite{suwajanakorn2017synthesizing} proposed a method belonging to the first category.
It relies on recurrent models (LSTMs) to predict person-specific 2D lip landmarks to guide 2D image generation.
Chung et al.~\cite{chung2017you} introduce a real-time approach for generating an RGB video of a talking face by directly mapping audio input to the video output space. 
Temporal generative adversarial networks (GANs) have been used for talking head generation~\cite{vougioukas2020realistic,zhou2020makelttalk}.
In particular, Vougioukas et al.~\cite{vougioukas2020realistic} present an approach for generating facial animation from a single RGB image using a temporal GAN.
While this approach directly utilizes speech information for talking head generation, MakeItTalk~\cite{zhou2020makelttalk} disentangles content from style and speaker identity, facilitating speech-driven generation that can be applied to diverse types of realistic and hand-drawn head portraits.
In the second category, an intermediate 3DMM~\cite{blanz1999morphable, egger20203d} is used to guide the 2D neural rendering of talking heads from audio \cite{thies2020nvp,song2022everybody,zhang2021flow,yi2020audio}, concentrating on the facial expressions.
Extending these approaches, Wang et al.~\cite{wang2021audio2head} add the head movements of the speaker to the synthesis.
Drawing inspiration from dynamic neural radiance fields~\cite{nerface}, several works~\cite{guo2021adnerf,yao2022dfa} leverage dynamic neural radiance fields to learn personalized talking head models that can be rendered under novel views, controlled by audio inputs.

\paragraph{Speech-Driven 3D Facial Animation}
Speech-driven 3D facial animation is a long-standing research question in computer graphics and animation.
Traditional methods are procedural techniques~\cite{de_martino_facial_2006,ezzat_miketalk:_1998, kalberer_face_2001, edwards2016jali} wherein the goal is to animate pre-defined facial rigs through procedural rules.
With the rise of deep learning, these methods have been extended by learning-based approaches~\cite{cao, gen-speech-animation, thies2020nvp, karras_audio-driven_2017, voca, meshtalk, faceformer, imitator}, where viseme patterns are directly learned from data.
A common theme for procedural techniques was to use Hierarchical Hidden Markov Models (HMM) as the basis for generating visemes from input text or audio, and subsequent facial animations were generated either through viseme-dependent co-articulation models~\cite{edwards2016jali, de_martino_facial_2006} or by blending facial templates~\cite{kalberer_face_2001}. 
Unlike these procedural approaches, data-driven approaches learn to generate 3D facial animation from data.
These approaches typically leverage pretrained speech models~\cite{deepspeech, wav2vec, wav2vec2.0} to generate an abstract and generalized representation of the input audio, which then serves as the input to a convolutional or auto-regressive model, mapping to either a 3D Morphable Model (3DMM) parameter space or directly to 3D meshes.
For instance, Karras et al.~\cite{karras_audio-driven_2017} exemplify learning a 3D facial animation model from small-scale but high-quality actor-specific 3D data.
While demonstrating a strong baseline, this method lacks generalization to new subjects and exhibits improper lip movements.
In VOCA~\cite{voca}, a model is trained on 3D data of multiple subjects, enabling the animation of corresponding identities from the input audio.
While this approach improves generalization over \cite{karras_audio-driven_2017}, the generalization remains limited as it requires one-hot encoding of identities at inference time.
MeshTalk~\cite{meshtalk} adopts a generalized approach by learning a categorical representation for facial expressions, auto-regressively sampling from this categorical space to animate a given 3D facial template mesh based on audio inputs.
In FaceFormer~\cite{faceformer} a pretrained Wav2Vec~\cite{wav2vec2.0} audio representation and a transformer-based decoder to regress displacements onto a template mesh is used.
Analogous to VOCA, FaceFormer incorporates a speaker identification code into the decoder, offering the flexibility to choose from talking styles present in the training set.
CodeTalker~\cite{xing2023codetalker} trains a Vector Quantized Variational Autoencoder (VQ-VAE) as motion prior.
Once the codebooks are trained, CodeTalker uses a transformer to learn the conditional distribution of codes given a speech signal.
More recently, EMOTE~\cite{EMOTE} utilizes the MEAD dataset~\cite{kaisiyuan2020mead} and generates pseudo-ground truth meshes using EMOCA~\cite{EMOCA_CVPR_2021}, a state-of-the-art 3D face reconstruction method.
Since MEAD contains different emotion and intensity labels, the trained model can generate speech-driven facial animation with various emotions, at the cost of slightly inferior lip-sync and realism compared to other approaches.

\paragraph{Motion Diffusion}
Diffusion models~\cite{sohl2015deep, ho2020denoising} have become a popular choice for generative tasks, among them the generation of images \cite{rombach2022high, dhariwal2021diffusion, gal2022textual, song2020denoising, salimans2022progressive}, videos \cite{blattmann2023videoldm, singer2022makeavideo, ho2022imagenvideo, zhou2023magicvideo}, 3D objects \cite{poole2022dreamfusion, chen2023fantasia3d, lin2023magic3d, yu2023textto3d}, audio \cite{kong2021diffwave, chen2020wavegrad}, and human motion \cite{tevet2023human, ma2022mofusion, yuan2023physdiff, zhang2022motiondiffuse, ren2023diffusion}. 
At their core, diffusion models are trained to iteratively denoise samples such that at inference time, new samples can be created from white-noise input. 
The stochastic nature of this process makes diffusion models highly suitable for modeling complex distributions.
In contrast to generative adversarial networks (GAN)\cite{dhariwal2021diffusion}, they show higher diversity and quality. 
Typically, the denoising process is conditioned on additional modalities such as text~\cite{rombach2022high}, audio~\cite{sun2023diffposetalk}, or depth maps~\cite{controlnet}. 
For an in-depth state-of-the-art report on diffusion models, we kindly refer the reader to the excellent surveys of Po et al.~\cite{po2023state} and Yang et al.~\cite{Yang2022DiffusionMA}.
In terms of methods that use diffusion for motion synthesis, MDM~\cite{mdm} and Mofusion~\cite{ma2022mofusion} are the closest to our work, producing impressive results for the task of body motion synthesis and editing from input text and music.
In contrast to both methods, we use a 1D-convolution architecture with concatenation-based condition injection to train on small-scale datasets more efficiently.

Despite the rapidly growing popularity of diffusion models, to the best of our knowledge, only two concurrent works apply them to the task of speech-driven face animation:
FaceDiffuser~\cite{FaceDiffuser_Stan_MIG2023} uses a pretrained speech representation model to convert the audio signal into sequences of latent feature vectors which condition a diffusion model that is based on recurrent GRU layers~\cite{chung2014empirical}. 
Similarly, DiffPoseTalk~\cite{sun2023diffposetalk} uses a diffusion model for speech-to-motion, however, they employ a transformer-based architecture and further propose a style encoder to personalize the synthesis based on short reference sequences. 
In contrast to these two methods, we use a convolution-based architecture which allows us to train our model on a small high-quality dataset and obtain higher synthesis quality and diversity. 
Furthermore, FaceDiffuser and DiffPoseTalk both perform auto-regressive inference, which restricts their applicability to animation editing tasks, e.g., motion between two keyframes cannot be synthesized consistently. 
Lastly, FaceDiffuser does not allow for personalization, and DiffPoseTalk's personalization capabilities are restricted within the space of the coefficients of a pretrained head model.
In contrast, our method regresses vertex displacements directly and allows for personalization with higher fidelity. 
\begin{figure*}[ht!]
    \centering
    \def\svgwidth{\linewidth}
\begingroup%
  \makeatletter%
  \fontsize{7}{12}
  \providecommand\color[2][]{%
    \errmessage{(Inkscape) Color is used for the text in Inkscape, but the package 'color.sty' is not loaded}%
    \renewcommand\color[2][]{}%
  }%
  \providecommand\transparent[1]{%
    \errmessage{(Inkscape) Transparency is used (non-zero) for the text in Inkscape, but the package 'transparent.sty' is not loaded}%
    \renewcommand\transparent[1]{}%
  }%
  \providecommand\rotatebox[2]{#2}%
  \newcommand*\fsize{\dimexpr\f@size pt\relax}%
  \newcommand*\lineheight[1]{\fontsize{\fsize}{#1\fsize}\selectfont}%
  \ifx\svgwidth\undefined%
    \setlength{\unitlength}{816bp}%
    \ifx\svgscale\undefined%
      \relax%
    \else%
      \setlength{\unitlength}{\unitlength * \real{\svgscale}}%
    \fi%
  \else%
    \setlength{\unitlength}{\svgwidth}%
  \fi%
  \global\let\svgwidth\undefined%
  \global\let\svgscale\undefined%
  \makeatother%
  \begin{picture}(1,0.39705882)%
    \lineheight{1}%
    \setlength\tabcolsep{0pt}%
    \put(0,0){\includegraphics[width=\unitlength,page=1]{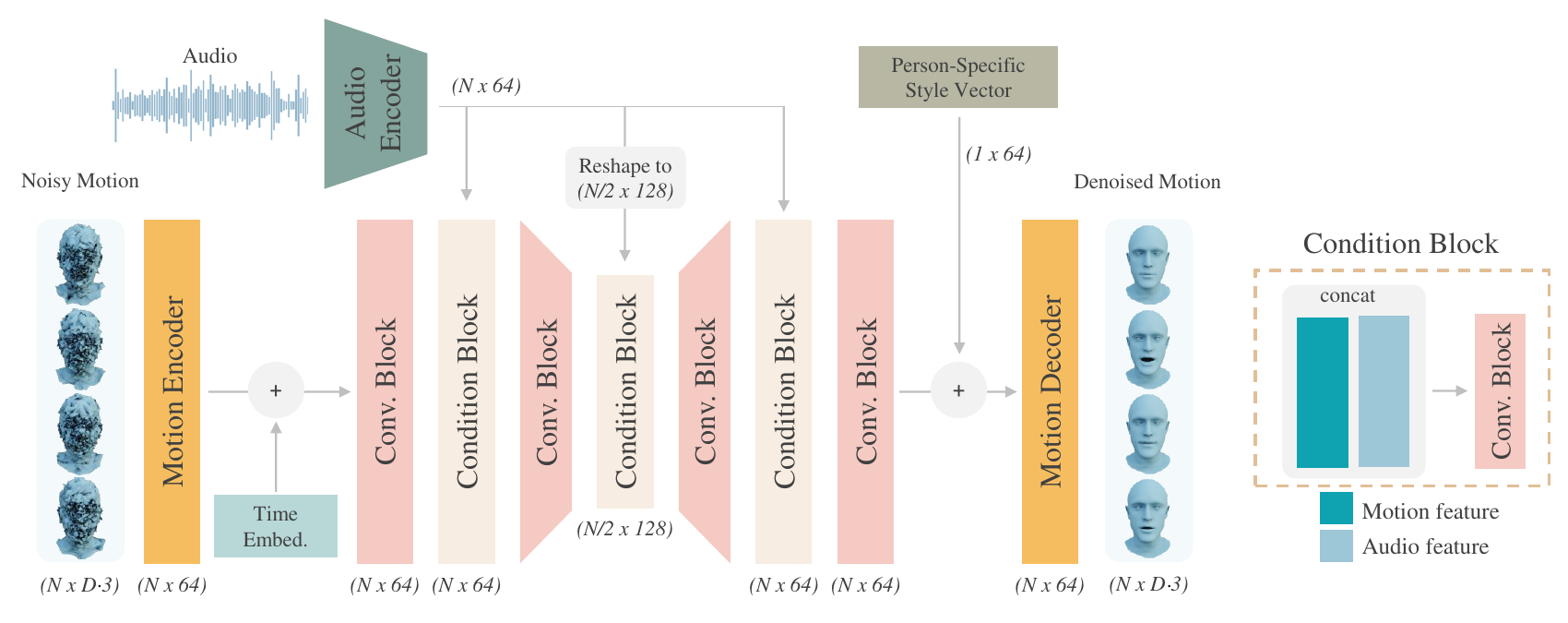}}%
    \put(0.045,0.26085784){\makebox(0,0)[lt]{\lineheight{1.25}\smash{\begin{tabular}[t]{l}$x_t$\end{tabular}}}}%
    \put(0.274,0.33767157){\makebox(0,0)[lt]{\lineheight{1.25}\smash{\begin{tabular}[t]{l}$\hat{A}$\end{tabular}}}}%
    \put(0.61609069,0.3107598){\makebox(0,0)[lt]{\lineheight{1.25}\smash{\begin{tabular}[t]{l}$S_i$\end{tabular}}}}%
    \put(0.725,0.26064951){\makebox(0,0)[lt]{\lineheight{1.25}\smash{\begin{tabular}[t]{l}$\hat{x}_0$\end{tabular}}}}%
  \end{picture}%
\endgroup%

      \caption{
Our method takes noised vertex displacements, denoted as $x_t$, and the diffusion time step embedding as inputs to predict a denoised sample $\hat{x}_0$, leveraging both the audio conditioning signal  $\hat{A}$ and a person-specific feature vector $S_i$.
Our approach employs wav2vec2.0~\cite{wav2vec2.0} for extracting audio features from the raw audio signal.
The audio condition is injected into the network by concatenation through a series of convolutional blocks.
Note that $N$ corresponds to the frame count of the sequence and $D$ to the number of vertices.
      }
      \label{fig:method}
\end{figure*}

\section{Preliminaries}
\label{sec:preliminaries}
\paragraph{Denoising Diffusion Probabilistic Models}
Our method is based on the diffusion framework of Sohl et al.~\cite{sohl2015deep}.
During the diffusion process, a data sample $x_0$ from the training distribution is iteratively disturbed by Gaussian noise for $T$ steps, resulting in a transition of the sample to white noise. 
The forward diffusion step $t$ is defined as: 
\begin{equation}
    x_t \sim q(x_t|x_{t-1}) = \mathcal{N}(x_t; \sqrt{ 1 - \beta_{t}} x_{t-1},  \beta_{t}I), t=1...T ,
\end{equation}
where $\beta_t$ is following a predefined variance schedule.
A denoising model is trained to reverse the diffusion process, hence to estimate $q(x_{t-1}|x_t)$.
Following recent work~\cite{mdm, sun2023diffposetalk}, we train a neural network $\theta$ to estimate $x_0$ from its noised version $x_t$: $\hat{x}_0 = \theta(x_t, t, C)$ with $C$ denoting additional conditions.
Following \cite{ho2020denoising}, the inverse diffusion process is then given through:
\begin{equation}
    q(x_{t-1}|x_t) = \mathcal{N}\left(x_{t-1}; \sqrt{\bar{\alpha}_{t-1}}\theta(x_t, t, C), (1-\bar{\alpha}_{t-1}) I\right) ,
\end{equation}
with $\alpha_t:=1-\beta_t$ and $\bar{\alpha}_t:=\prod_{k=1}^t\alpha_k$.
For generating new samples, we randomly sample $x_T$ from a Gaussian distribution and iteratively denoise it until $t=0$ is reached.

To add diversity, we employ Classifier-Free Guidance (CFG)~\cite{ho2022classifierfree} and calculate the output as a weighted sum of the conditional and unconditional prediction:
\begin{equation}
    \theta_s(x_t, t, C) := \theta(x_t, t, \emptyset) + s \cdot \left[\theta(x_t, t, C)-\theta(x_t, t, \emptyset)\right] ,
\end{equation}
where $s$ is the guidance scale and $\theta(x_t, t, \emptyset)$ denotes the unconditional prediction in which we set the audio conditions to zero.
Note that while CFG is typically used with a guidance scale $>1$ to enhance alignment with the condition, we set it to values $<1$ (0.5 unless specified otherwise) to increase diversity.

\paragraph{Audio Encoding}
Similar to other state-of-the-art methods~\cite{voca, faceformer, imitator, codetalker}, we adopt the pretrained Wav2Vec2.0~\cite{wav2vec2.0} model to generate audio features from the raw audio signal. 
Wav2Vec2.0 uses a self-supervised learning approach to map audio to quantized feature vectors with $768$ channels. 
We resample the output of Wav2Vec2.0 via linear interpolation to match the sampling rate of the motion sequences (30fps for VOCAset~\cite{voca}).
A trainable linear layer is applied to project the feature vectors to $64$ channels, resulting in a speech representation $\hat{A} \in \mathbb{R}^{N\times64}$ for $N$ frames.
%


\section{Method}
\label{sec:method}
Our goal is to synthesize and edit facial animations given speech audio as a conditioning input. 
We represent facial animations as a sequence of 3D vertex displacements that can be applied on top of a template mesh.
To generate those displacements from audio, we employ a diffusion-based model that is trained to iteratively denoise the displacement sequences.
This architecture does not only produce stochastic outputs, it also allows us to edit the animation sequence by defining keyframing.
As facial motions are person-specific, we design the architecture such that it can be adapted and fine-tuned to specific subjects, only requiring a short video sequence of the actor.
In \Cref{fig:method}, we show an overview of our method, which is detailed in the following. 
As written above, we represent facial animations as sequences of 3D vertex displacements w.r.t. a template mesh.
Let $x_0\in \mathbb{R}^{N\times D \cdot 3}$ denote such a sequence where $N$ is the sequence length and $D$ is the number of vertices in the template mesh. 
The input to our diffusion model $\theta$ is a noised vertex displacement sequence $x_t\in\mathbb{R}^{N\times D \cdot 3}$ and we aim to predict its noise-free counterpart: $\hat{x}_0=\theta(x_t, t, C)$ given diffusion step $t$ and conditions $C$. 
As a first step, we employ a single fully connected layer as \textit{Motion Encoder} to project $x_t$ to a 64-dimensional latent space.
We positionally encode the diffusion step $t$ \cite{sohl2015deep}, map it to the latent space with a linear layer, and add it to the encoded $x_t$. 
We apply a series of 1D-convolution blocks to first reduce the temporal dimension of the activations, followed by an upsampling convolution block to restore the original temporal dimension.
Each convolution block is followed by a condition block to incorporate the audio features.
The condition blocks concatenate the input features with the audio conditions and apply a dimension-preserving convolution. 
We add a person-specific feature vector $S_i\in\mathbb{R}^{1\times64}$ to the output of the convolutional layers prior to applying to the \textit{Motion Decoder}.
This produces the final noise-free sample $\hat{x}_0$. 
Similar to the \textit{Motion Encoder}, the \textit{Motion Decoder} is a single fully connected layer.
Note that in our formulation, the condition $C$ represents the set of both the per-frame audio features $\hat{A}$ and the person-specific feature vector $S_i$.
In contrast to state-of-the-art methods on 3D facial animation synthesis that use transformer architectures~\cite{faceformer, imitator, codetalker, sun2023diffposetalk}, we take inspiration from Pavllo et al.~\cite{pavllo20193d} and adopt a 1D-convolutional network as our backbone. 
Specifically, instead of infusing the condition through an attention mechanism, we use feature concatenation. 
We found that these architecture changes are crucial for efficiently training the model on the small available VOCA training set~\cite{voca} (see \Cref{tab:ablation}). 
Note that while other methods train transformer architectures on bigger datasets with pseudo-ground-truth annotations, we show that our architecture changes allow us to achieve superior results while training on a smaller, yet high-quality dataset. 
In particular, the fully convolutional nature of our network allows us to randomly crop the sequences to $30$ frames for training and generalize to sequences of arbitary length at inference time. 
We find this data augmentation strategy to be vital for improved generalization and convergence in the unconditional setting.
Note that this data augmentation strategy is not possible for transformer-based architectures since they rely on a consistent positional encoding, which prevents them from generalizing to longer sequences.
While auto-regressive motion synthesis could in theory resolve this limitation, it would make crucial animation editing tasks impossible, such as inbetweening distant motion frames. 
Further, we empirically find that for the unconditional case, in which the audio conditions are set to $0$, transformer-based architectures do not converge on the small VOCA training set due to its limited size. 
However, as outlined in \Cref{sec:preliminaries}, unconditional synthesis is crucial for synthesis diversity.  

\subsection{Training}
Similar to \cite{tevet2023human, sun2023diffposetalk}, we train our diffusion model to predict the ground truth vertex displacements $x_0$ from their noised counterparts $x_t$: 
\begin{equation}
 \mathcal{L}_{\text{simple}} = || x_{0} - \theta(x_{t},t,C) || ^2 .
\end{equation}
In comparison to predicting the applied noise which is common practice in related work~\cite{ma2022mofusion, zhang2022motiondiffuse, rombach2022high}, we empirically found that predicting the ground truth displacements yields better convergence in the unconditional and person-specific fine-tuning setup. 
Furthermore, we take inspiration from \cite{imitator, voca} and add a velocity loss $\mathcal{L}_\text{vel}$ to improve temporal smoothness:
\begin{equation}
 \mathcal{L}_{\text{vel}} = \dfrac{1}{N-1} \sum_{n=1}^{N} || (x_{0,n} - x_{0, n-1}) - (\hat{x}_{0,n} - \hat{x}_{0,n-1}) || ^2 ,
\end{equation}
where $x_{0,n}$ denotes the ground truth vertex displacements in frame $n$.
Our final training objective is formulated as:
\begin{equation}
    \label{eq:total}
 \mathcal{L}_{\text{total}} = \mathcal{L}_{\text{simple}}  + \lambda_{\text{vel}}\cdot\mathcal{L}_{\text{vel}} .
\end{equation}
We empirically set $\lambda_{\text{vel}}=10.0$ unless specified otherwise. 
Note that during training, we randomly set the audio condition $C$ to 0 in $10\%$ of the cases in order to enable unconditional synthesis at inference time.


\subsection{Person-Specific Fine-tuning}
For capturing the speaking style of a subject that is not part of the training set, we require a short reference talking head video. The facial movements are extracted with the state-of-the-art monocular face tracker MICA~\cite{mica}. 
We use the tracked meshes as pseudo ground truth and fine-tune the entire model to fit the expression distribution of the target subject using the training objective from Eq.~\eqref{eq:total}.
\section{Dataset}
\label{sec:data}
We train our model on the VOCAset~\cite{voca} since it provides high-quality, speech-aligned 3D face scan sequences. 
It consists of $12$ actors ($6$ female and $6$ male) with $40$ sequences each with a length of $3$-$5$ seconds, resampled at $30$fps.
Following previous work \cite{imitator}, we use the train/val/test set split of $8,2,2$ actors.
All $40$ sequences of the training actors are used during training. 
However, for the test and validation, only $20$ sequences without overlap with the speech scripts of the training sequences are used.
For the style adaption experiment, we split the $40$ sequences of the test actors to $18, 2, 20$ for train/val/test sets. 
The test sequences of the experiments w/ and w/o style adaptation are identical, allowing a direct comparison of the scores in \Cref{tab:exprs_quan_study}.

We evaluate person-specific fine-tuning for in-the-wild sequences on the video sequences from  Imitator~\cite{imitator}. The provided videos are 2 minutes long which we divide into 60/30/30 seconds for train/val/test respectively.
\section{Results}
\label{sec:results}

\begin{figure}
    \begin{center}
        \input{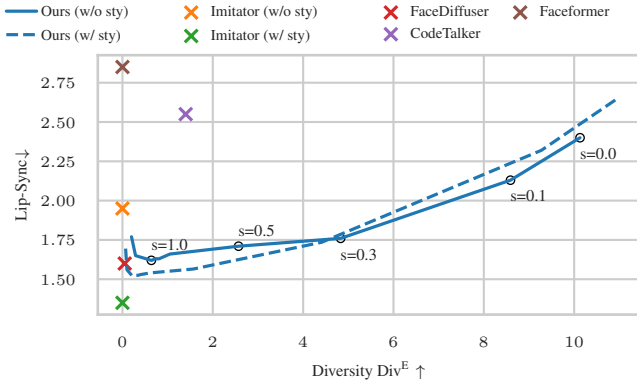}
    \end{center}
    \vspace{-0.5cm}
    \caption{
    We investigate the impact of the classifier-free-guidance scale $s$~\cite{ho2022classifierfree} using the 'Lip-sync' and $Div^E$ metrics.
    Lower guidance values yield animations with significantly more diverse motion but inferior lip-sync quality. Conversely, higher guidance values result in high-quality animation with reduced diversity. 
    For our experiments, we maintain a fixed guidance value of $0.5$, a sweet spot that balances high diversity with excellent lip-sync quality.
    }
     \label{fig:abl_div}
\end{figure}

\begin{figure}[t!]
    \centering
    \def\svgwidth{\linewidth}
    \input{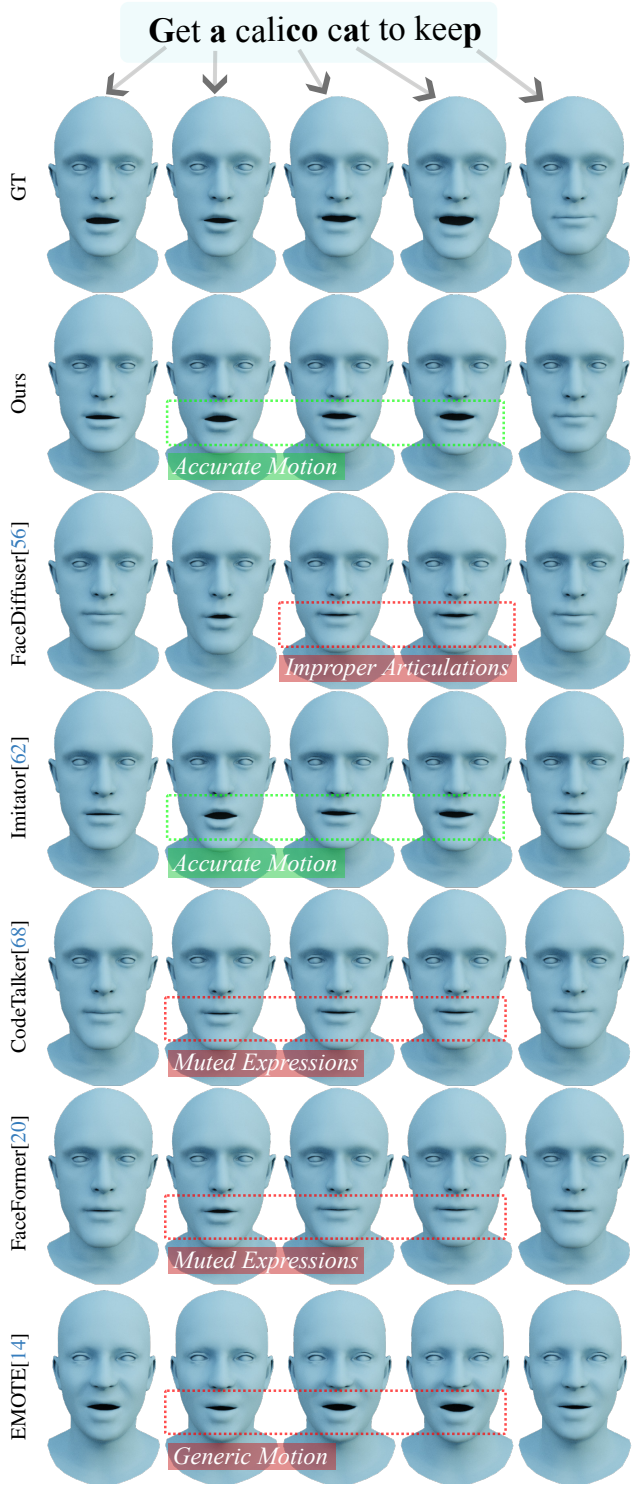}
      \caption{
      Qualitative comparison. 
        Only our method and Imitator produce expressive motions that match the target speaking style. 
        While Imitator synthesizes similarly convincing animations, its outputs are not diverse (see \Cref{tab:exprs_quan_study}) and it cannot be used for animation editing. 
        \vspace{-.5cm}
        }
      \label{fig:baseline_comp}
\end{figure}

We evaluate and we compare our method against state-of-the-art methods: VOCA~\cite{voca}, Faceformer~\cite{faceformer}, CodeTalker~\cite{codetalker}, EMOTE~\cite{EMOTE} , FaceDiffuser~\cite{FaceDiffuser_Stan_MIG2023} and Imitator~\cite{imitator}.
\Cref{fig:baseline_comp} shows a qualitative comparison to the baselines on a test sequence from the VOCAset. 
We find that only our method and Imitator produce expressive facial animations that match the speaking style of the target subject. 
For additional qualitative results, we refer to the suppl. video.

\paragraph{Quantitative Comparison} 
In \Cref{tab:exprs_quan_study}, we present a quantitative evaluation based on the following metrics: 
\textit{Lip-Sync} measures the lip synchronization using Dynamic Time Warping to compute the temporal similarity~\cite{imitator}.
\textit{Lip-max}~\cite{meshtalk} reports the mean of the maximal per-frame lip distances.
${L_2^\text{lip}}$ and ${L_2^\text{face}}$ correspond to the mean $L_2$ vertex errors for the lip region and the entire face respectively.
Additionally, we adopt the diversity metric $Div^E$ proposed by Ren et al.~\cite{ren2023diffusion} to assess the diversity of animations generated from the same audio.
Note that only Imitator~\cite{imitator} and our method allow for optimizing person-specific speaking styles.
For a fair comparison, we report results both with and without person-specific style optimization for these two models. 
For the non-personalized synthesis, we find that our method with the default guidance scale $s=0.5$, improves synthesis diversity by over 80\% compared to the closest competitor (see row 4 and 6 of \Cref{tab:exprs_quan_study}). 
Furthermore, the same model performs second-best in terms of \textit{Lip-Sync} and is competitive on all other metrics. 
This demonstrates that our method with guidance scale $s=0.5$ offers significantly improved synthesis diversity while still ensuring plausible lip synchronization. 
Note that we can use the guidance scale parameter to freely trade synthesis diversity for lip-sync accuracy. 
As we increase the guidance scale to $s=1.0$ (row 7 of \Cref{tab:exprs_quan_study}), we are able to match the \textit{Lip-Sync} score of the top-performing baseline.
We visualize the trade-off between lip-sync accuracy and synthesis diversity in \Cref{fig:abl_div}.
We find that the guidance scale $s$ is an effective tool to increase synthesis diversity beyond all baselines with only a small loss of lip-sync accuracy for $0.3\leq s \leq 1.0$.
When personalizing our model, we find that while the synthesis diversity decreases, all other scores improve. The resulting model now consistently outperforms all non-personalized models on all metrics. 
Note that we only require $\sim100s$ of video to personalize our method to an unseen identity. 
The moderate decline in synthesis diversity during personalization is an expected behavior and even is an indicator of successful personalization. 
During personalization, the model learns to suppress movements that do not align with the target identity and as a natural consequence, the synthesis diversity is reduced.

In comparison to Imitator~\cite{imitator} after personalization, we achieve higher synthesis diversity and comparable accuracy scores, yet we are not able to outperform this baseline. 
However, note that Imitator is a deterministic model that does not allow for stochastic synthesis. Also, in contrast to Imitator and all other baselines, our method is the only one that enables animation editing like motion inbetweening.

\begin{table}[t]
    \resizebox{\linewidth}{!}{%
        \begin{tabular}{cl|ccccc} \toprule
             &\textbf{Method} & $\mathbf{Div^{E}}$ $\uparrow$  & \textbf{Lip-Sync} $\downarrow$ & \textbf{Lip-max} $\downarrow$  & $\mathbf{L_2^{lip}}$ $\downarrow$ & $\mathbf{L_2^{face}}$ $\downarrow$ \\ \midrule
                 & & & \multicolumn{3}{c}{Non-Personalized Synthesis} \\ 
                \cmidrule(r){3-7} 
                1 & VOCA~\cite{voca} & $-$ & $5.30$ &  $7.06$ &  $0.20$ & $0.94$ \\
                2 & Faceformer~\cite{faceformer} & $-$ & $2.85$ &  $5.41$ &  $0.14$ & $\mathbf{0.80}$ \\
                3 & Imitator~\cite{imitator} & $-$ & $1.95$ &  $\mathbf{4.95}$ &  $\mathbf{0.12}$ & $0.85$ \\
                4 & CodeTalker~\cite{codetalker} & $1.40$  & $2.55$ &  $5.02$ &  $0.14$ & $0.88$ \\
                5 & FaceDiffuser~\cite{FaceDiffuser_Stan_MIG2023} & $0.05$ &  $\mathbf{1.60}$ &  $5.20$ &  $0.16$ & $0.89$  \\
                  6 & Ours$_{s=0.5}$ (w/o sty) & $\mathbf{2.57}$ & $1.71$ &  $5.20$ &  $0.15$ & $0.86$  \\
                   7 & Ours$_{s=1.0}$ (w/o sty) & $0.64$ & $1.62$ &  $5.13$ &  $0.15$ & $0.84$  \\
                \midrule
                 & & & \multicolumn{3}{c}{Personalized Synthesis} \\ 
                \cmidrule(r){3-7} 
                8 & Imitator (w/ sty) & $-$ & $\mathbf{1.35}$ &  $\mathbf{3.43}$ &  $\mathbf{0.09}$ & $\mathbf{0.76}$ \\
                9 & Ours (w/ sty) & $\mathbf{1.57}$ & $1.56$ &  $4.01$ &  $0.11$ & $0.78$  \\
                \midrule
        \end{tabular}
    }
    \caption{The quantitative results from the VOCAset~\cite{voca} demonstrate that our method outperforms the baseline in generating diverse motions. 
    It achieves comparable performance to the state-of-the-art regression method in terms of lip-sync accuracy, both in personalized and non-personalized setups.
    }
    \label{tab:exprs_quan_study}
\end{table}

\begin{figure}[t!]
    \centering
    \includegraphics[width=\linewidth]{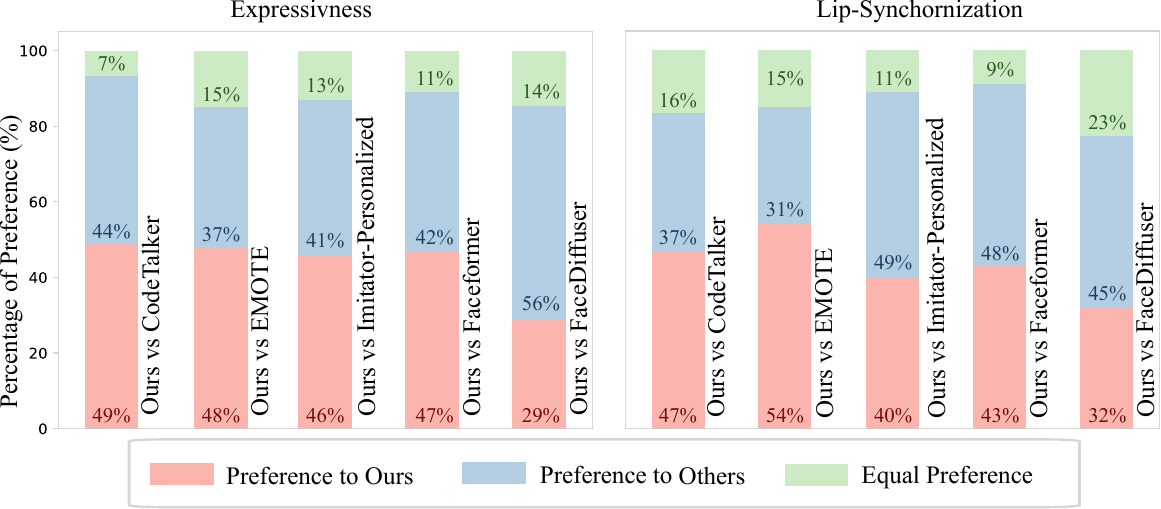}
      \caption{
     User-study results on the VOCAset~\cite{voca}. Overall, our performance matches or surpasses the state-of-the-art (SOTA) in terms of expressiveness and lip-sync, except for FaceDiffuser~\cite{FaceDiffuser_Stan_MIG2023}.
     FaceDiffuser generates animations with less diversity and high lip-sync.
     Additionally, we evaluate the person-specific speaking-style modeling by comparing against Imitator-Personalized~\cite{imitator}. Users were presented with a reference video and asked to select the synthesized method that closely resembled it. Notably, $40\%$ of users favored our method for better style-similarity, while $17\%$ found both methods equally comparable.
      }
      \label{fig:user_study}
\end{figure}

\paragraph{User Study}
We conducted an A/B user study to assess our method's perceptual performance. 
We sample 20 sequences combined from the VOCAset test set and the in-the-wild sequences from Imitator, resulting in 100 A/B comparisons across five baselines.
On Amazon Mechanical Turk(AMT), we divided the A/B comparisons into $5$ HITs~(Human Intelligence Task), each with $25$ individual assignments. For each HIT, users select their preference for a method based on expressiveness and lip-synchronization.
The results in ~\Cref{fig:user_study} show that our method outperforms most of the baselines in terms of lip-synchronization and expressiveness, except for the concurrent work FaceDiffuser.
The difference in performance w.r.t. FaceDiffuser is an inherent trade-off between diversity and lip-sync quality.
Our method with a guidance value of $1.0$ matches the performance of FaceDiffuser with better diversity and additionally, our method allows for person-specific fine-tuning and motion editing.
We, further, conduct a second user study to evaluate the speaking style preservation of our personalized model in comparison to Imitator. 
To this end, the AMT users rated the similarity based on a reference video and the synthesized videos of the VOCA test set.  
40\% of the users preferred our method, 43\% preferred Imitator, 17\% voted for tie. 
This demonstrates that our method for the first time allows to synthesize diverse personalized face animations and enables animation editing without significantly reducing speaking style faithfulness. 

\begin{figure}[t!]
    \centering
    \includegraphics[width=\linewidth]{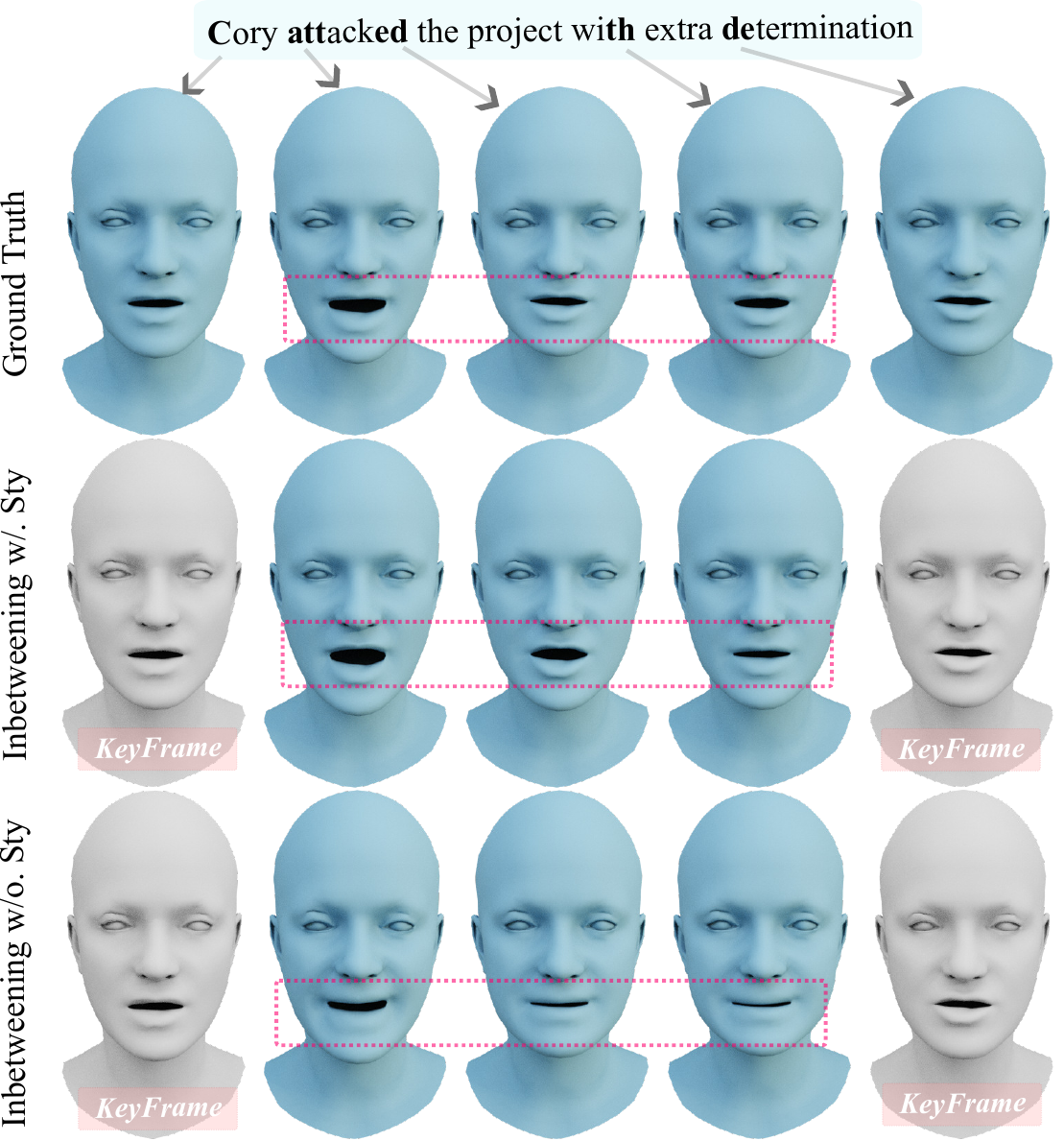}
      \caption{
      Qualitative evaluation of the importance of person-specific finetuning for motion editing. 
      As highlighted in \textcolor{purple}{purple}, without finetuning we observe an abrupt change in speaking style between the keyframes and the generated motion, thus rendering the results unrealistic.
      }
      \label{fig:keyframing_sty_adap}
\end{figure}

\paragraph{Motion Editing}
We show motion editing using keyframes in \Cref{fig:keyframing_sty_adap}.
In this application, we selectively replace the predicted denoised vertex-displacement sequences $\hat{x}_0$ with ground truth values during the denoising process.
This is similar in spirit to well-established diffusion-based image inpainting methods~\cite{lugmayr2022repaint}.
We additionally show unconditional motion synthesis and editing results in the supplemental material.
As can be seen in \Cref{fig:keyframing_sty_adap}, the personalization of the motion synthesis is important to match the talking style, preventing an abrupt style change.

\paragraph{Ablation}
We evaluate the benefits of our 1D-convolutional architecture over the transformer-based architectures used in the baseline methods, and the attention-based convolutional approach in MoFusion~\cite{ma2022mofusion}.
To this end, we compare our proposed architecture against two variants. (i) We replace the conditional convolution blocks with attention layers such as those used in Mofusion (row 1 in \Cref{tab:ablation}). (ii) We replace the entire backbone with the transformer-based architecture from Faceformer (row 2 in \Cref{tab:ablation}).
We observe that both changes to our method significantly worsen all scores. 
This confirms the improved effectiveness of our proposed architecture when training on small datasets (see discussion in \Cref{sec:method}). 

\textit{Training data for person-specific fine-tuning:}
We evaluate the impact of the data set size for person-specific fine-tuning in \Cref{tab:ablation} row 4-7. To this end, we perform fine-tuning on the VOCAset test set by varying the dataset size to $5$/$30$/$60$/$100$s.
We observe that while the fine-tuning on $5$ s diverges, $30$s and $60$s suffice to achieve acceptable results. 
For $100$s of data, our model is able to synthesize motion with a better \textit{Lip-sync} and diversity $Div^E$.

\begin{table}[t!]
    \resizebox{\linewidth}{!}{%
        \begin{tabular}{cl|ccccc} \toprule
             &\textbf{Method} & $\mathbf{Div^{E}}$ $\uparrow$ & \textbf{Lip-Sync} $\downarrow$ & \textbf{Lip-max} $\downarrow$  & $\mathbf{L_2^{lip}}$ $\downarrow$ & $\mathbf{L_2^{face}}$ $\downarrow$ \\ \midrule
               & & & \multicolumn{3}{c}{(a) Architecture ablation} \\ 
                \cmidrule(r){3-7} 
                1 & Ours (w/ conv attn) & $0$ & $1.68$ &  $5.17$ &  $0.15$ & $0.88$  \\
                2 & Ours (w/ FF arch) & 0 &  $3.49$ &  $6.2$ & $0.19$ & $0.96$ \\
                3 & Ours & $1.57$  & $1.56$ &  $4.01$ &  $0.11$ & $0.78$\\
                \midrule
                & & & \multicolumn{3}{c}{(b) Person-specific Fine-tuning} \\ 
                \cmidrule(r){3-7} 
                4 & Ours ($\sim$ 5s) & $29.95$ & $4.89$ &  $13.05$ &  $0.37$ & $1.95$ \\
                5 & Ours ($\sim$ 30s) & $0.18$  & $1.81$ &  $4.90$ &  $0.13$ & $0.86$\\
                6 & Ours ($\sim$ 60s) & $0.67$  & $1.69$ &  $4.18$ &  $0.12$ & $0.83$\\
                7 & Ours ($\sim$100s) & $1.57$  & $1.56$ &  $4.01$ &  $0.11$ & $0.78$\\
               \midrule
                 & & & \multicolumn{3}{c}{(c) Audio noise ablation} \\ 
                \cmidrule(r){3-7} 
                8 & Ours (high noise)   & $6.41$ & $2.56$ &  $6.68$ &  $0.19$ & $0.94$    \\
                9 & Ours (med. noise) & $2.54$ & $1.97$ &  $4.93$ &  $0.13$ & $0.84$  \\
                10 & Ours (low noise)    & $1.85$& $1.78$ &  $4.4$ &  $0.12$ & $0.77$      \\
            \bottomrule
        \end{tabular}
    }
    \caption{
    We ablate our method with respect to (a) architecture, (b) person-specific fine-tuning dataset size, and (c) audio noise level robustness. 
    Our architecture outperforms attention-based conditioning mechanisms (row 1), and the transformer backbone of Faceformer~\cite{faceformer} (row 2). 
    Further, we show that $30$s of video suffice to perform person-specific fine-tuning while $100$s further improve all scores (row 4-7). 
    Our method is robust wrt. medium and low audio noise levels (row 9-10). 
    }
    \label{tab:ablation}
\end{table}

\textit{Sensitivity study:}
Additionally, we conducted a noise sensitivity experiment similar to ~\cite{voca,imitator}, where we added white noise to the input audio with a negative gain of 36db (low), 24db (medium), and 12db (high).
As reported in ~\Cref{tab:ablation} rows 8-10, our method produces robust high-quality facial animations for low and medium noise levels.

\section{Discussion}
\label{sec:discussion}

Our proposed method excels in synthesizing and editing diverse 3D facial animations based on speech.
Its architecture is carefully designed for efficient training on a small, yet high-quality dataset. 
Nonetheless, we believe the performance can be further improved by increasing the amount of training data. 
In the user study, our method outperforms most baselines and is on par with the concurrent work FaceDiffuser~\cite{FaceDiffuser_Stan_MIG2023}.
At the same time, our method synthesizes animations with higher diversity and allows the editing of existing animations such as inserting new motion frames.
Currently, we do not consider head motion (e.g. neck rotation) in our method, due to the lack of this data in the VOCAset.
However, we believe that it can be extended to head motion, given a suitable dataset.
%

\section{Conclusion}
\label{sec:conclusion}

With 3DiFACE we present the first method that can both generate and edit diverse 3D facial animations from speech input.
Employing classifier-free guidance provides us with an effective tool to balance synthesis diversity and accuracy allowing us to generate animations with unprecedented diversity while outperforming or matching all baselines with respect to synthesis accuracy. 
Through personalization, we can extract person-specific speaking styles from short ($\sim 100s$) videos which significantly improves performance. 
%
Further, our architecture allows us to edit animations by using keyframes. 
We are convinced that these properties make 3DiFACE a powerful tool for content creators and are excited for future applications. 

%
%
 
\section{Acknowledgements}
This project has received funding from the Mesh Labs, Microsoft, Cambridge, UK.
The authors thank the International Max Planck Research School for Intelligent Systems (IMPRS-IS) for supporting Balamurugan Thambiraja.
We would like to thank Malte Prinzler for the support in writing and rendering figures.
We also want to thank Berna Kabadayi, Peter Kulits, Ikhsanul Habibie  and Prasanna Mayilvahanan for valuable discussion and feedback.
\begin{figure}[ht!]
    \centering
    \includegraphics[width=\linewidth]{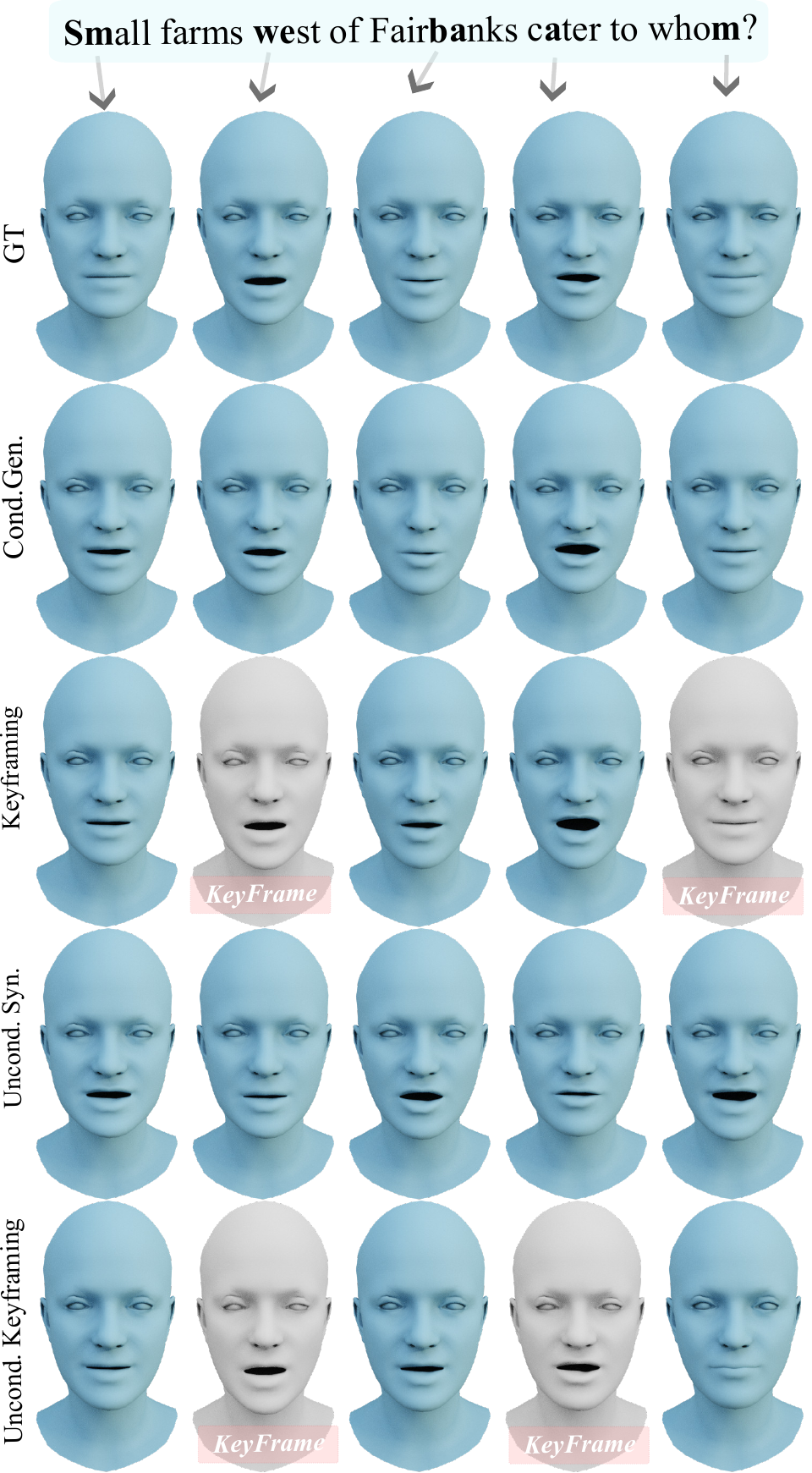}
      \caption{
      Qualitative illustration of motion inbetweening using our conditional and unconditional model. 
      In row 2 and 3, we showcase a sequence synthesized conditionally and subsequently refined using keyframes.
      In Row 4 (Uncond. Syn.), we present our unconditional synthesis results. As observed from the results, our model can unconditionally synthesize facial animations that appear plausible.
      Further in row 5 (Uncond. Keyframing), we see that our method can unconditionally inbetween facial animation while preserving the speaking-style of the target actor.
      This progression demonstrates our model's capabilities: from conditional synthesis and keyframe-based editing to unconditional synthesis and editing, while maintaining the actor's speaking-style.
      }
      \label{fig:ablation_main}
\end{figure}

\section{Additional Applications}

\subsection{Unconditional motion synthesis and editing:} 
While unconditional motion synthesis has been extensively applied in the motion synthesis domain~\cite{tevet2023human, raab2022modi}, to the best of our knowledge, its application in 3D facial animation synthesis remains widely unexplored.
The significance of an unconstrained facial motion synthesis method cannot be overstated. It holds substantial potential for various applications, such as animating background characters like NPCs in movies and games. 
Additionally, it enables targeted editing of specific facial elements—such as eye blinks and eyebrow motions—since these non-verbal facial expressions often exhibit weak or no correlation with audio features.
Moreover, an unconditional model serves as a valuable motion prior for downstream tasks, extending its utility beyond synthesis and editing applications. 
Our demonstration of unconditional synthesis and editing are showcased in~\Cref{fig:ablation_main}, underscoring the potential and versatility of such unconstrained models for 3D facial animation synthesis.

\begin{table}[ht!]
    \resizebox{\linewidth}{!}{%
        \begin{tabular}{cl|ccccc} \toprule
             &\textbf{Method} & $\mathbf{Div^{E}}$ $\uparrow$  & \textbf{Lip-Sync} $\downarrow$ & \textbf{Lip-max} $\downarrow$  & $\mathbf{L_2^{lip}}$ $\downarrow$ & $\mathbf{L_2^{face}}$ $\downarrow$ \\ \midrule
                1 & Ours (synthesis) & $1.35$ & $1.4$ &  $3.41$ &  $0.09$ & $0.69$  \\
                \midrule
                2 & Ours (Ip 5\%)  & $1.27$ & $1.17$ &  $3.19$ &  $0.09$ & $0.65$   \\
                3 & Ours (Ip 10\%) & $1.24$  & $1.15$ &  $2.98$ &  $0.08$ & $0.61$  \\
                4 & Ours (Ip 20\%) & $1.15$  & $1.01$ &  $2.66$ &  $0.07$ & $0.54$  \\
                5 & Ours (Ip 50\%) & $0.9$   & $0.68$ &  $1.74$ &  $0.05$ & $0.33$ \\
                \midrule
                6 & Ours (1KF/sec) & $1.26$ & $1.28$ &  $3.17$ &  $0.09$ & $0.65$ \\
                7  & Ours (2KF/sec) & $1.14$ & $1.2$ &  $2.98$ &  $0.08$ & $0.62$ \\
                8 & Ours (3KF/sec) & $1.05$ & $1.1$ &  $2.77$ &  $0.07$ & $0.59$  \\
            \bottomrule
        \end{tabular}
    }%
    \caption{We quantitatively evaluate our motion editing capability on all the test sequences of the subject 024 in the VOCAset~\cite{voca}.
    To this end, first we preserve 5\%, 10\%, 20\%, 50\% of the starting and ending frames, and then perform inbetweening for the intermediate motion sequences. 
    In addition, we assess the robustness of the inbetweening by randomly inserting keyframes (KF): 1KF/sec, 2KF/sec, and 3KF/sec.
    From the metrics, we can see that the synthesis quality increases significantly with adding more keyframes, which is a clear indication that the model matches the ground truth and produces realistic motion.
    For animators and artists, this means that they can insert any number of keyframes they want and have fine-grained control over the motion synthesis.
    Note that keyframes could also stem from previously generated motion sequences using our method (iterative refinement).
    }
    \label{tab:motion_quan_study}
\end{table}

\subsection{Motion Keyframing/Inbetweeing:}
We evaluate the performance of motion inbetweening with respect to the input data.
To this end, we preserve  5\%, 10\%, 20\%, 50\% of the starting and ending frames, and then perform inbetweening for the intermediate motion sequences. 
Furthermore, we assess the robustness of the inbetweening by randomly inserting keyframes at different rates: 1KF/sec, 2KF/sec, and 3KF/sec. 
These evaluations are conducted for all sequences of the test subject 024 from the VOCAset~\cite{voca}, and the resulting metrics are presented in \Cref{tab:motion_quan_study}.
These metrics demonstrate the efficacy of our method in robustly editing motion.

\section{Implementation:}

\subsection{Baselines:}
For VOCA~\cite{voca}, Faceformer~\cite{faceformer}, Imitator~\cite{imitator} and FaceDiffuser~\cite{FaceDiffuser_Stan_MIG2023}, we use the pre-trained model provided in the offical repositories.
For CodeTalker~\cite{codetalker}, we adapt the official implementation to add the functionality of generating diverse motion. Especially, we re-train the audio conditioned codebook sampling (stage 02) to randomly sample a code from top 'm' closest codes instead of always using the closest code.
This process is in spirit close to training the language-based models, where a new diverse text sequence is generated by sampling the 2nd or 3rd closest language token over the token with maximum probability.
By adapting this method, we ensure that CodeTalker could generate diverse samples for a given audio input.
For EMOTE~\cite{EMOTE}, we request the authors to run their method on the VOCAset~\cite{voca} and use it for the qualitative and perceptual user-study.

\subsection{Training Details:}

We train our method using ADAM~\cite{kingma2017adam} with a learning rate of \textit{1e-4} for 140K iterations with a batch size of $64$.
Our diffusion framework is based on the Gaussian diffusion from Nichol \textit{et al.}~\cite{nichol2021improved}, we set the diffusion step to $500$ for our experiments.
During training, we randomly crop the sequences to length of $30$ frames.
Our light-weight architecture enables to train our model on a single Nvidia quadro P6000 $32GB$ within $30$ hours.
The lightweight architecture is also critical for person-specific style-adaption with a short reference video.
For person-specific speaking-style, we use the same training setup as from the generalized setting, except that we only train it for $30K$ iterations.
For evaluating the best checkpoint, we fix the guidance scale $s=0.99$ and evaluate all the saved checkpoints on the validation set.
Further, we fix the best checkpoint and vary the guidance scale from \textit{s=0, 0.1 ... to 1.0} with increment of $0.1$ and find the best guidance factor.
From our experiment, we found the guidance scale of $0.5$ balances the lip-synchornization and diversity and provides best results.

\subsection{Inference}
Our method takes $3.15$sec to produce $1$sec ($30$ frame) of facial animation on a single Nvidia GeForce RTX 3090 $24GB$, compared to $5.78$sec for the concurrent method FaceDiffuser~\cite{FaceDiffuser_Stan_MIG2023}.

\section{Broader Impact}
\label{sec:broader_impact}
We introduce a method for realistic facial animation synthesis and editing that matches the speaking-style of any given target actor.
These animations hold promise for driving virtual avatars in AR or VR settings, especially, in immersive communication technologies.
Yet, it is essential to acknowledge the potential pitfalls of such advancements, notably in the realm of 'DeepFakes.' 
By employing voice cloning techniques, our method can generate 3D facial animations that drive digital avatar methods like~\cite{Gafni_2021_CVPR, Zielonka2022InstantVH,grassal2022neural, kabadayi24ganavatar, bharadwaj2023flare}, which could be abused for identity theft, cyberbullying, and various criminal activities.
Advocating for transparent research practices, we strive to illuminate the risks associated with technology misuse. 
Sharing our implementation aims to foster research in digital multimedia forensics, particularly in developing synthesis methods crucial for training data utilized in spotting forgeries~\cite{roessler2019faceforensics++}.
%
{
    \small
    \bibliographystyle{utils/ieeenat_fullname}
    \bibliography{main}
}

\end{document}